%% file: main.tex
\documentclass[runningheads]{llncs}
\usepackage{amsmath,amssymb, amsfonts, stmaryrd}   
\usepackage{wrapfig}
\usepackage{newclude}
\usepackage{bm}
\usepackage{booktabs}
\usepackage{array} 
\usepackage{xcolor}
\usepackage{pifont}
\usepackage{graphicx}
\usepackage{colortbl}
\newcommand{\sj}[1]{{\color{red}{[SJ:#1]}}}

\begin{document}
\title{CMViM: Contrastive Masked Vim Autoencoder for 3D Multi-modal Representation Learning for AD classification}
%
%
\author{Guangqian Yang\inst{1} \and
Kangrui Du\inst{2}\and
Zhihan Yang\inst{3} \and Ye Du \inst{1} \and  Yongping Zheng \inst{1,4} \and Shujun Wang \inst{1,4} }
\authorrunning{G. Yang et al.}
%
\institute{Department of Biomedical Engineering, The Hong Kong Polytechnic University, Hong Kong
SAR, China
\and School of Computer Science and Engineering, University of Electronic Science and Technology of China, Chengdu, China \and School of Mechanical, Electrical and Information Engineering, Shandong University, Weihai, China \and Research Institute for Smart Ageing, The Hong Kong Polytechnic University, Hong Kong SAR, China \\
\email{shu-jun.wang@polyu.edu.hk}}
%
\maketitle              
%
\begin{abstract}
Alzheimer's disease (AD) is an incurable neurodegenerative condition leading to cognitive and functional deterioration. Given the lack of a cure, prompt and precise AD diagnosis is vital, a complex process dependent on multiple factors and multi-modal data.
While successful efforts have been made to integrate multi-modal representation learning into medical datasets, scant attention has been given to 3D medical images.
In this paper, we propose Contrastive Masked Vim Autoencoder (CMViM), 
the first efficient representation learning method
tailored for 3D multi-modal data.
Our proposed framework is built on a masked Vim autoencoder to learn a unified multi-modal representation and long-dependencies contained in 3D medical images. We also introduce an intra-modal contrastive learning module to enhance the capability of the multi-modal Vim encoder for modeling the discriminative features in the same modality, and an inter-modal contrastive learning module to alleviate misaligned representation among modalities. 
Our framework consists of two main steps:
1) incorporate the Vision Mamba (Vim) into the mask autoencoder to reconstruct 3D masked multi-modal data efficiently. 
2) align the multi-modal representations with contrastive learning mechanisms from both intra-modal and inter-modal aspects. 
Our framework is pre-trained and validated ADNI2 dataset and validated on the downstream task for AD classification. The proposed CMViM yields 2.7\% AUC performance improvement compared with other state-of-the-art methods.

\end{abstract}

\input{sections/1-intro}
\input{sections/2-method}

\input{sections/3-exp}
\input{sections/4-con}

%
%
%
\bibliographystyle{splncs04}
\bibliography{main}

\end{document}

%% file: sections/1-intro.tex
\section{Introduction}


Alzheimer's disease (AD) has been widely known as a progressive and irreversible neurodegenerative disease in the elderly and the main cause of dementia \cite{scheltens2021alzheimer,joe2019cognitive}, which leads to incurable memory loss, cognitive impairment, and functional deficits with loss of functional independence \cite{canonici2012functional}. 
As the global population continues to age, the prevalence of AD is prone to rise significantly, resulting in a substantial burden on healthcare systems and society as a whole \cite{zvvevrova2018alzheimer}. 
%
Therefore, early diagnosis of AD is crucial for timely prevention and intervention to alleviate the progressive neurodegeneration and above society burden \cite{rasmussen2019alzheimer}. 
However, the diagnosis of AD is a complex procedure which does not rely on a single factor but relies on {multi-modal data}~\cite{aviles2022multi}, like cognitive and neurological tests, brain scans, psychiatric evaluation and so on\footnote{\url{https://www.nia.nih.gov/health/alzheimers-and-dementia/what-dementia-symptoms-types-and-diagnosis}}.

In the medical scenario, the huge potential of multi-modal representation learning has been harnessed, yielding significant improvements through pretraining on extensive data and subsequent adaptation for downstream tasks 
\cite{chen2023contrastive,wang2023ecamp,wei2023masked}.
However, acquiring medical data can be challenging and often results in small dataset sizes. In such cases, representation learning or pre-training can be highly beneficial. However, such representation learning strategies for 3D multi-modal data, such as AD-related tasks, are seldom discussed. Hence, it is imperative to investigate the strategy of Representation Learning for 3D multi-modal data analysis.

Contemporary representation learning, like masked autoencoding (MAE) \cite{he2022masked}, which is primarily grounded on foundation models, predominantly utilizes a single type of sequence model: the Transformer \cite{vaswani2017attention} and its integral attention layer \cite{bahdanau2014neural}. However, these architectures are limited in their ability to model anything beyond a finite window and exhibit quadratic complexity \cite{gu2023mamba}, which is not optimal for high-resolution 3D medical images with a significant number of voxels. Recently, Mamba-based models \cite{ma2024u,liu2024vmamba,xing2024segmamba,zhu2024vision,guo2024mambamorph} have proven their efficiency and effectiveness in long-sequence modeling, potentially emerging as the solution for long-range dependency modeling in visual tasks. However, most Mamba-based models are initially trained from scratch, leaving the effect of pretraining on these models in 3D multi-modal data analysis tasks unclear. Comprehending the efficacy of pretraining Mamba-based models equipped with representation learning could provide crucial insights for numerous tasks.

The prevalent methodologies for representation learning currently rely on contrastive learning~\cite{chen2020simple,he2020momentum}, MAE~\cite{zhou2023advancing}, and their combination~\cite{chen2023contrastive}. MAE mainly focuses on learning local detailed information for image reconstruction, making it less efficient to learn discriminative representation. However, contrastive learning can model relations and differences among images and modalities. Therefore, the mask autoencoder and contrastive learning could potentially complement each other. For instance, Chen \textit{et.al.}~\cite{chen2023contrastive} proposed contrastive masked image-text modeling utilized by recovering masked image patches and text reports through MAE and contrastive learning.
It utilized the representation alignment of different modality data which is the paired masked images and reports. Nevertheless, such cross-modal representation alignment is not enough, which neglects the importance of learning inter-modal discriminated information related to downstream tasks, especially for AD classification.

In this paper, we introduce the first efficient representation learning method for 3D multi-modal data,
\textbf{Contrastive Masked Vim autoencoder}(CMViM).
Our framework is designed with two steps:
1) incorporate the Vision Mamba (ViM) into the mask autoencoder to reconstruct 3D masked multi-modal data, to effectively model the long-range dependencies inherent in 3D medical data. 
2) align the multi-modal representations using contrastive learning mechanisms from both intra-modal and inter-modal perspectives to enhance the discriminative capabilities of image representations. \textbf{Intra-modal Contrastive Learning} is performed on the data from the same modality, the positive pairs are from the original model and a momentum-updated model during each iteration of MAE pre-training. 
While \textbf{Inter-modal Contrastive Learning} is implemented among different modalities to align cross-modality representations.
Our framework is pre-trained and validated on the collected ADNI2 dataset for AD classification. The proposed CMViM yields at least 2.7\% AUC improvement compared with the state-of-the-art representation learning methods. 
%
%


%% file: sections/2-method.tex
\section{Method}
\begin{figure}[t]
\centerline{\includegraphics[width=1\linewidth]{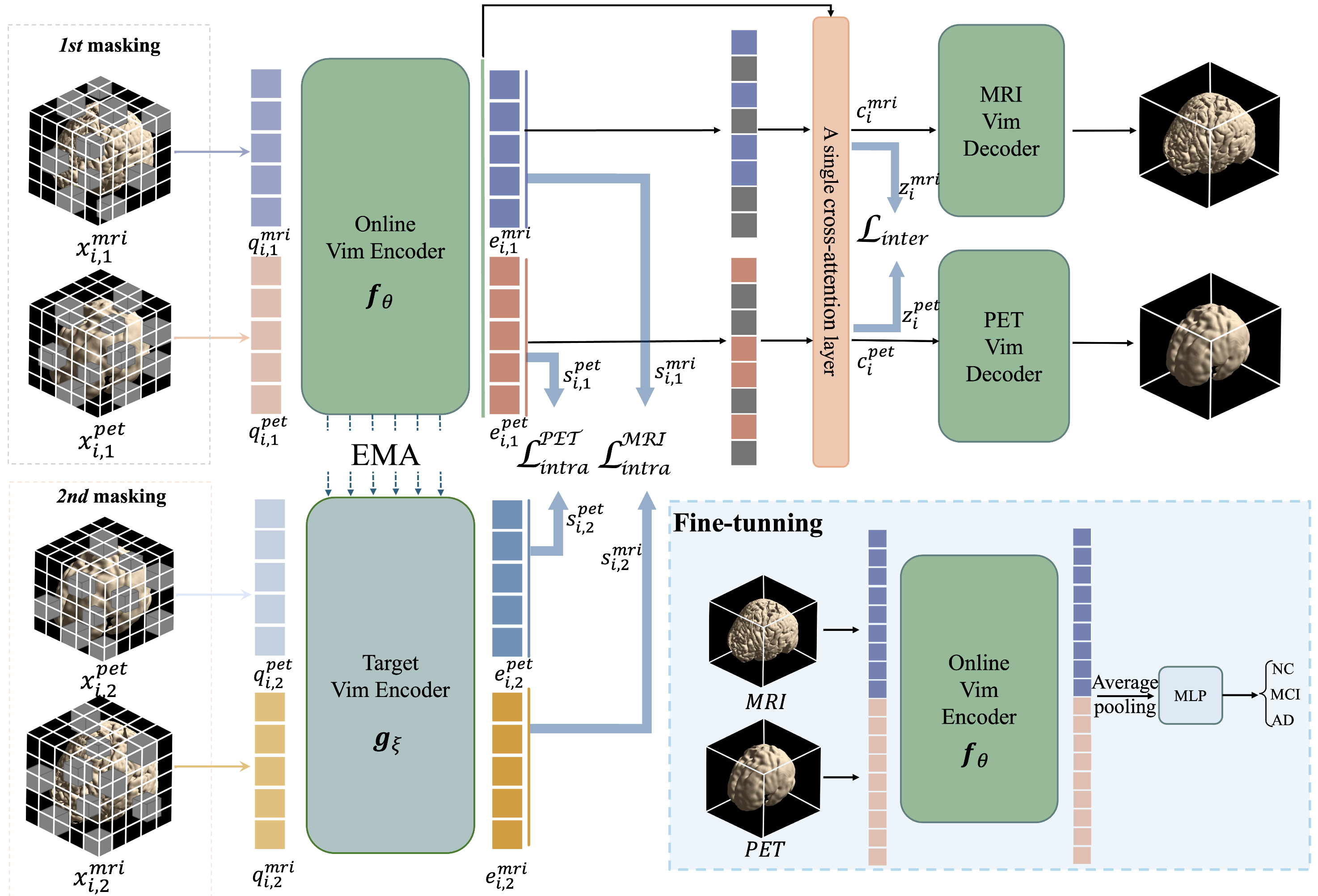}}
\caption{Overview of our proposed Contrastive Masked Vim Autoencoder (CMViM). For pre-training, we build a multi-modality masked autoencoder based on Vim blocks for 3D medical visual representation learning. To further strengthen the abilities to capture disease-related features and relieve the representation misalignment between different modalities, we respectively introduce an intra-modal contrastive learning module and an inter-modal contrastive learning module. Then, the multi-modality Vim encoder pre-trained by our proposed CMViM is finetuned to AD classification.}
\label{fig}
\end{figure}

Our proposed method CMViM is designed for the representation learning for 3D multi-modal data analysis. The overall architecture is depicted in Fig.~\ref{fig} 
In this section, we initially provide a brief overview of Vision Mamba. Subsequently, we delve into the specifics of our multi-modality masked Vision Mamba for 3D representation learning. Lastly, we offer detailed descriptions of our inter-modal and intra-modal contrastive learning modules.

\subsection{Vision Mamba}

Mamba \cite{gu2023mamba} has demonstrated state-of-the-art performance and efficiency in NLP \cite{pioro2024moe, anthony2024blackmamba}.
Recently, Zhu \textit{et al.} \cite{zhu2024vision} proposed Vim, by applying Mamba into a supervised visual representation learning field.

Vim first transform the 2D image $\bm{x} \in \mathbb{R}^{H\times W \times C}$ into the flattened patches $\{\bm{x}_j \in \mathbb{R}^{ P^2 \times C}\}_{j=1}^J$, where $(H, W)$ is the image size, $C$ is the number of channels, $P$ is the size of image patches and $J$ is the total number of patches.
Then, it linearly projects each patch to the vector with size $D$ and adds a class token $\bm{t}_{cls}$ to represent the whole patch sequence:
\begin{equation}
    \begin{aligned}
        \bm{T}_0 = [ \bm{t}_{cls};\ \bm{x}_0\bm{W};\ ... ;\  \bm{x}_J\bm{W} ] + \bm{E}_{pos},
    \end{aligned}
\end{equation}
where $\bm{W}$ is the learnable projection layer and $ \bm{E}_{pos}$ is the position embedding.
%
The sequences are then fed into the Vim encoder with a residual connection:
\begin{equation}
    \bm{T}_l = \text{Vim}(\bm{T}_{l-1}) + \bm{T}_{l-1},
\end{equation}
where each layer is applied in a bidirectional manner.
Then, the class token $\bm{T}^0_L$ from the last layer $L$ is normalized and input into a multi-layer perceptron (MLP) head to get the prediction $\hat{p}$:
\begin{equation}
    \hat{p} = \text{MLP}(\text{Norm}(\bm{T}^0_L)).
\end{equation}
It is worth noting that Vim outperforms the standard ViT in both accuracy and efficiency when compared to models with equivalent parameter counts.
To facilitate the long-range dependency modeling ability for 3D multi-modal medical data, we extend the 2D mamba into the 3D version and utilize it as the backbone of Masked Autoencoder to form Masked Vim Autoencoder.

\subsection{Multi-modal Masked Vim autoencoder for 3D Representation Learning}
%
In this subsection, we introduce the Multi-modal Masked Vim Autoencoder, for feature extraction and feature interaction between multi-modality data (T1-MRI and PET images).
As shown in Fig.~\ref{fig}, the Multi-modal Masked Vim Autoencoder contains a Vim encoder for encoding projected tokens of unmasked patches to latent representations and two modality-specific Vim decoders for reconstructing input images. We will elaborate on the encoder and decoder separately below.

\vspace{-0.2cm}
\subsubsection{Encoder}
Considering a multi-modality dataset with N pairs of T1-MRI and PET 3D images, denoted as $\mathcal D = \{ \bm{x}_{i}^{mri}, \bm{x}_{i}^{pet}\}_{i=1}^{N}$. 
We first divide the original T1 and PET 3D images into non-overlapping 8$\times$8$\times$8 patches.
Subsequently, these 3D patches were projected into tokens using modality-specific linear projection layer $\{P_{mri},P_{pet}\}$. 
Then, we added two learnable position embeddings to projected tokens from T1 and PET images. 
Subsequently, We employ a random masking strategy to mask out a large portion of 3D patch embeddings.
In this study, we adopt a masking ratio of 75\% following \cite{he2022masked}.
The remaining visible patch embeddings were concatenated to a sequence of patch embeddings and then input into an encoder consisting of 12 vision mamba layers for capturing long-dependencies, producing two encoded tokens $\{ \bm{e}_{i}^{mri}, \bm{e}_{i}^{pet}\}_{i=1}^{N}$.

\vspace{-0.2cm}
\subsubsection{Decoder}
The decoder is designed for multi-modality data reconstruction.
Specifically, two separated shallow decoders with a single Vim layer were employed for different modalities with the same architecture but unshared weights. To start with, the encoded tokens $\{ \bm{e}_{i}^{mri}, \bm{e}_{i}^{pet}\}_{i=1}^{N}$ output from the encoder were concatenated with a set of learnable tokens corresponding to masked patch embeddings. 
To integrate cross-modality information into learnable tokens, a single cross-attention layer that views encoded single-modality tokens as queries and all tokens as keys and values following MultiMAE \cite{bachmann2022multimae}. 
Finally, modality-specific tokens $\{ \bm{c}_{i}^{mri}, \bm{c}_{i}^{pet}\}_{i=1}^{N}$ output by a single cross-attention layer were input into a lightweight decoder with a vision mamba layer followed by an MLP layer to reconstruct the pixels. The multi-modality mamba for 3D image reconstruction was supervised by two mean square error loss $\mathcal L_{rec}^{mri}$ and $\mathcal L_{rec}^{pet}$ to compare the difference reconstructed pixels and the original pixels for T1 and PET images respectively.

\subsection{Multi-modal Contrastive Learning}
To enhance the ability of the multi-modal Vision Mamba encoder for learning discriminative disease-related representations, we propose an intra-modal contrastive learning module to encourage our multi-modal mamba encoder to learn similar and dissimilar visual representations for the same modality. Meanwhile, to relieve the misalignment between T1-MRI and PET images, we use an inter-modal contrastive learning module to reduce the representation difference from paired T1-MRI and PET images.

\vspace{-0.2cm}
\subsubsection{Intra-modal Contrastive Learning Module}
The first step of contrastive learning is to construct positive sample pairs and negative sample pairs. For MAE, it is reasonable that MAE is supposed to generate similar visual representations for two sequences of unmasked embedded tokens from the same image after randomly masking two times, generating distinctive features for different images of the same modality. The module built two encoders with the same framework, named as online encoder $\bm{f}_\theta$ and target encoder $\bm{g}_\xi$. We aim to train the online encoder to effectively capture intra-modal discriminative features. The network parameter $\xi$ of target encoder $\bm{g}_\xi$ is updated by using exponential moving average (EMA) of parameter $\theta$ of online encoder $\bm{f}_\theta$, calculated as:
\begin{equation}
    \xi = \beta\times\xi + (1-\beta)\times\theta
\end{equation}
where $\beta\in [0,1]$ is a momentum coefficient. Following \cite{he2020momentum}, $\beta$ is fixed to 0.999 in the following implementation.

For a single modality, considering two sequences of unmasked embedded tokens $\{\bm{q}_{i,1},\bm{q}_{i,2}\}_{i=1}^{B}$ from the same image, B is batch size. The two sequences are individually fed into the online encoder $\bm{f}_\theta$ and target encoder $\bm{g}_\xi$, generating two sequences of encoded tokens $\{\bm{e}_{i,1}, \bm{e}_{i,2}\}_{i=1}^{B}$. Then we apply a global average pooling layer and a projection layer to $\{\bm{e}_{i,1}\}_{i=1}^N$ and $\{\bm{e}_{i,2}\}_{i=1}^B$. Therefore, InfoNCE loss (Eq.(7)) is adopted to calculate inter-modal contrastive loss.

\begin{equation}
    \{\bm{e}_{i,1}, \bm{e}_{i,2}\}_{i=1}^{B} = \{\bm{f}_\theta(\bm{q}_{i,1}), \bm{g}_\xi(\bm{q}_{i,2})\}_{i=1}^{B}
\end{equation}
\begin{equation}
    \{\bm{s}_{i,1}, \bm{s}_{i,2}\}_{i=1}^{B} = \{Proj_1(Avgpool(\bm{e}_{i,1})), Proj_2(Avgpool(\bm{e}_{i,2}))\}_{i=1}^{B}
\end{equation}
\begin{equation}
\mathcal L_{intra}^{mri/pet} = -\frac{1}{2B}\sum_{i=1}^B\left[\log\frac{\exp(\bm{s}_{i,1}\bm{s}_{i,2}^{T})}{\sum \exp(\bm{s}_{i,1}\bm{s}_{i,2}^{T})}+ \log\frac{\exp(\bm{s}_{i,2}\bm{s}_{i,1}^{T})}{\sum \exp(\bm{s}_{i,2}\bm{s}_{i,1}^{T})}\right]
\end{equation}
\subsubsection{Inter-modal Contrastive Learning Module}
The encoded tokens output by the cross-attention layer in the decoder, denoted as $\{\bm{c}_{i}^{mri},\bm{c}_{i}^{pet}\}_{i=1}^{B}$, are input into a global average pooling layer and a projection layer, generating  $\{\bm{z}_{i}^{mri},\bm{z}_{i}^{pet}\}_{i=1}^{B}$. InfoNCE loss using in inter-modal contrastive learning can be represented as follows:
\begin{equation}
\mathcal L_{inter} = -\frac{1}{2B}\sum_{i=1}^B\left[\log\frac{\exp(\bm{z}_{i}^{mri}{\bm{z}_{i}^{pet}}^{T})}{\sum \exp(\bm{z}_{i}^{mri}{\bm{z}_{i}^{pet}}^{T})}+ \log\frac{\exp(\bm{z}_{i}^{pet}{\bm{z}_{i}^{mri}}^{T})}{\sum \exp(\bm{z}_{i}^{pet}{\bm{z}_{i}^{mri}}^{T})}\right]
\end{equation}

\subsection{Training Procedure and Experimental Details}

\subsubsection{Pre-training}
We employ a similar two-stage training strategy to CMITM \cite{chen2023contrastive} due to the difficulty of simultaneously optimizing reconstruction loss and inter-modal contrastive learning loss, which was observed in the previous works \cite{chen2023contrastive,jiang2023layer}.
In the first stage, the framework was trained under the supervision of reconstruction loss $\mathcal L_{res}$ and intra-modal contrastive loss $\mathcal L_{intra}^{mri/pet}$:
\begin{equation}
    \mathcal L_{1st} = \mathcal L_{res}^{mri} + \mathcal L_{res}^{pet} + \alpha(L_{intra}^{mri}+L_{intra}^{pet})
\end{equation}
where $\alpha$ is a hyper-parameter, and $\alpha$ is empirically set to 0.005. The framework was trained for 1600 epochs with a batch size of 48, using an AdamW optimizer \cite{loshchilov2017decoupled} with a cosine decaying learning rate and a weight decay of 0.05. The initial learning rate is set to 0.005.
In the second stage, the inter-modal contrastive loss was combined into an overall loss (Eq. (10)) to train the framework for 500 epochs using an initial learning rate of 0.0005 and a $\beta$ of 0.2:
\begin{equation}
    \mathcal L_{2nd} = \mathcal L_{res}^{mri} + \mathcal L_{res}^{pet} + \alpha(L_{intra}^{mri}+L_{intra}^{pet}) + \beta\mathcal L_{inter} 
\end{equation}

\subsubsection{Finetuning}
The pre-trained multi-modal Vim encoder was finetuned to AD classification (NC/MCI/AD), which was trained for 300 epochs using AdamW optimizer with an initial learning rate of 0.001 and a weight decay of 0.05, under the supervision of a focal loss with a $\gamma$ of 3.

%% file: sections/3-exp.tex
\section{Experimental Results}


\subsection{Dataset Description}
This study was conducted on the Alzheimer’s Disease Neuroimaging Initiative (ADNI) dataset. ADNI is a landmark research project that aims to provide a better understanding of AD through the collection and analysis of comprehensive data. It was initially launched in 2004. we collected 1292 samples with paired 3D T1-MRI and PET images from ADNI2. The collected dataset contains three categories, Normal Control (NC), Mild Cognitive Impairment (MCI), and Alzheimer's disease (AD). 

\subsection{Data Pre-processing and Dataset division}

\subsubsection{Pre-processing Pipeline} Our data pre-processing process can be divided into three steps: 1) We randomly sample some subjects and register their PET images to the corresponding T1 images using \texttt{mri\_robust\_register} in FreeSurfer \cite{fischl2012freesurfer}, from which the best one is chosen as the template. 2) All T1 and PET images are then registered to this template. 3) SynthStrip \cite{hoopes2022synthstrip} is finally applied to perform skull stripping for all images. After data normalization, all T1-MRI and PET images are resized into 64$\times$64$\times$64.
\vspace{-0.3cm}
\subsubsection{Dataset Division} We randomly divide the collected dataset into a training set (70\%), a validation set (10\%), and a test set (20\%). we pre-train our multi-modal Vision Mamba encoder using the training set and the validation set. To achieve AD classification, we first load pre-trained weights to train our Multi-modal Vim Encoder on the training set, then optimize hyper-parameters on the validation set, and eventually evaluate the model performance for AD classification on the test set.
\vspace{-0.3cm}
\subsection{Comparison with State-of-the-art Method}

We compare our proposed CMViM with a multi-modal self-supervised framework (MultiMAE~\cite{bachmann2022multimae}), several state-of-the-art methods for AD classification. MultiMAE is a novel multi-modal self-supervised framework based on pure ViT for 2D natural image analysis, we modify it into a 3D version for AD classification based on paired T1-MRI and PET images. Other works, M3T~\cite{jang2022m3t}, MedicalNet~\cite{chen2019med3d}, 3D ResNet-50~\cite{hara3dcnns}, and nnMamba~\cite{gong2024nnmamba} were originally designed to process single-modal data. For a fair comparison, we concatenate T1-MRI and PET 3D images, forming a multi-modal input with an input channel of 2. The concatenated input containing multi-modal information was input into those methods for AD classification.

\begin{table}[h]
\centering
\setlength{\tabcolsep}{4pt}
\renewcommand{\arraystretch}{1.2}
\caption{Quantitative comparison with state-of-the-art method for multi-modal supervised learning and AD classification. \dag: The latest version pre-trained on 23 datasets.}
\label{tab:comparison}
\scalebox{0.87}{
\begin{tabular}{l|lccc|ccc}
\toprule
Method                    & Backbone  & \#Param. & Pre-train Set & \#Sample & Acc  & AUC  & F1-score \\ \midrule
\rowcolor{gray!20} \multicolumn{8}{l}{\textit{\textbf{Trained from scratch:}}} \\
M3T~\cite{jang2022m3t}                       &      CNN+ViT     &    7M    &       -        &        -         &   59.8   & 75.5     &     56.0     \\
Hara et al.~\cite{hara3dcnns}.             & ResNet-50 &    46M    &       -        &          -       &   64.5   &   82.9   &   63.5       \\
nnMamba~\cite{gong2024nnmamba}                  &   CNN+Mamba     &     13M    &       -        &         -        &    65.2  &  81.3    &    62.7      \\
\rowcolor{gray!20} \multicolumn{8}{l}{\textit{\textbf{Pre-trained:}}} \\ 

MedicalNet~\cite{chen2019med3d}                & ResNet-50 &    46M    &     3DSeg-23$^\dag$      &       $>$1638               &     68.1 &  81.4    & 65.8         \\


MultiMAE~\cite{bachmann2022multimae}                  &    ViT        &   89M     &      ADNI2         &        1292           & 59.4 & 79.2 & 59.0     \\

\textbf{CMViM (ours)}              & Vim       &    50M    &            ADNI2   &              1292    & \textbf{69.3} & \textbf{84.1} & \textbf{66.0}     \\ \bottomrule
\end{tabular}}
\end{table}

Table. \ref{tab:comparison} shows our proposed CMViM outperforms MultiMAE and other state-of-the-art methods for AD classification. Especially for comparison with MultiMAE, CMViM not only achieves an extraordinary improvement in model performance but also dramatically reduces the number of network parameters, which demonstrates its impressive abilities for modeling long-range global information for 3D representation learning with excellent computational efficiency.

\begin{table}[h]
\centering
\setlength{\tabcolsep}{6pt}
\renewcommand{\arraystretch}{1.2}
\caption{Ablation study on our proposed method.}
\scalebox{0.86}{
\begin{tabular}{l|cc|ccc|ccc}
\toprule
& \multicolumn{2}{c|}{Architecture} & \multicolumn{3}{c|}{SSL Method} & \multicolumn{3}{c}{Evaluation Metric} \\ \midrule
& Vim            & ViT           & MAE    & InterCL    & IntraCL   &  Acc   &  AUC  &  F1-score        \\ \midrule
 1&      \ding{52}          &               &        &            &           &  60.6   &  79.4   &58.6\\
 2&                &   \ding{52}    &   \ding{52}     &            &           &   59.4&79.2&59.0   \\
 3&     \ding{52}           &               &     \ding{52}   &            &           &   67.7&83.7&64.7      \\
 4&       \ding{52}         &               &   \ding{52}     &        \ding{52}        &       &  68.9&84.0&65.1      \\
  \rowcolor{gray!20}  5& \ding{52}               &               &      \ding{52}  &      \ding{52}     &   \ding{52}        &   69.3&84.1&66.0 \\
                 \bottomrule
\end{tabular}}
\end{table}
\vspace{-0.4cm}
\subsection{Ablation Study}
Finally, we conducted an ablation study to validate the effectiveness of key components using our proposed CMViM framework, as shown in Table 2. The results show that replacing the ViT layers with Vim layers can remarkably improve the model performance. Moreover, Multi-modality Vim without SSL even achieves a slightly better performance than Multi-modality ViT fine-tuned on pre-trained weights from MAE. Besides, we also conclude that combining two classical paradigms, MAE and contrastive learning, can significantly enhance 3D multi-modal representation learning for Vim.

%% file: sections/4-con.tex
\section{Conclusion}

In this study, we introduce a novel CMViM for the first 3D medical representation learning in multi-modal data analysis. Our method utilizes a Vision Mamba encoder to address model redundancy, short-sequence modeling, and computational inefficiency in modern ViT-based networks. We incorporate an intra-modal contrastive learning module to enhance representation learning and an inter-modal contrastive module to mitigate representation misalignment among different modalities. Experimental results on the AD diagnosis dataset demonstrate that our method outperforms other state-of-the-art methods in AD diagnosis.